 \documentclass[11pt,a4paper]{article}
\usepackage{hyperref}
\usepackage[utf8]{inputenc}
\usepackage{subcaption}
\usepackage{multirow}
\usepackage{bm}
\usepackage{verbatim}

\usepackage{color}
\usepackage[left=3cm, right=3cm, bottom=4cm, top=4cm]{geometry}

\usepackage[table,xcdraw]{xcolor}
\usepackage{booktabs}
\usepackage[numbers]{natbib}
\usepackage{graphicx}
\usepackage{amssymb}
\usepackage{amsmath}
\usepackage{mathabx}
\usepackage{soul}

\usepackage{hhline}

\usepackage{epstopdf}

\usepackage{enumitem}
\usepackage[font=small,skip=2pt]{caption}
\graphicspath{ {Figures/}}

\newcolumntype{L}[1]{>{\raggedright\let\newline\\\arraybackslash\hspace{0pt}}m{#1}}
\newcolumntype{C}[1]{>{\centering\let\newline\\\arraybackslash\hspace{0pt}}m{#1}}
\newcolumntype{R}[1]{>{\raggedleft\let\newline\\\arraybackslash\hspace{0pt}}m{#1}}

\newcommand{\ttt}{\boldsymbol{\theta}}

\usepackage{ifthen}
\makeatletter
\renewcommand\@cite[2]{%
Ref.~#1\ifthenelse{\boolean{@tempswa}}
{, \nolinebreak[3] #2}{}
}
\renewcommand\@biblabel[1]{#1.}

\newcommand{\bblue}[1]{#1}
\newcommand{\rred}[1]{#1}

\makeatother
\newcounter{cntcomment}

\makeatletter
\def\ps@pprintTitle{%
\let\@oddhead\@empty
\let\@evenhead\@empty
\def\@oddfoot{}%
\let\@evenfoot\@oddfoot}
\makeatother

\title{Constrained-CNN Losses for Weakly Supervised Segmentation}
\author{Hoel Kervadec*\footnote{Corresponding author: hoel.kervadec.1@etsmtl.net}, Jose Dolz*, Meng Tang**\\Eric Granger*, Yuri Boykov**, Ismail Ben Ayed*\\\small{*\'ETS Montréal, QC, Canada}\\\small{**Department of computer science, University of Waterloo, ON, Canada}}
\date{}

\begin{document}
    \maketitle
    % \begin{frontmatter}

        % \author[LIVIA]{Hoel Kervadec\corref{mycorrespondingauthor}}
        % \author[LIVIA]{Jose Dolz}
        % \author[Western]{Meng Tang}
        % \author[LIVIA]{Eric Granger}
        % \author[Western]{Yuri Boykov}
        % \author[LIVIA]{Ismail Ben Ayed}

        % \address[LIVIA]{\textsc{Livia}, \'ETS Montréal, QC, Canada}

        % \address[Western]{Department of computer science, University of Waterloo, ON, Canada}

        % \cortext[mycorrespondingauthor]{Corresponding author: hoel.kervadec.1@etsmtl.net}

    \begin{abstract}
        Weakly-supervised learning based on, e.g., partially labelled images or image-tags, is currently attracting significant attention in CNN segmentation as it can mitigate the need for full and laborious pixel/voxel annotations.
        %Weak supervision, e.g., in the form of partial labels or image tags, is currently attracting significant attention in CNN segmentation as it can mitigate the lack of full and laborious pixel/voxel annotations.
        %, a common problem in medical imaging.
        Enforcing high-order (global) inequality constraints on the network output (for instance, to constrain the size of the target region) can leverage unlabeled data, guiding the training process with domain-specific knowledge.
        % Enforcing high-order (global) inequality constraints on the network output (for instance, on the size of the target region) can leverage unlabeled data, guiding training with domain-specific knowledge.
        Inequality constraints are very flexible because they do not assume exact prior knowledge. However, constrained Lagrangian dual optimization has been largely avoided in deep networks, mainly for computational tractability reasons. To the best of our knowledge, the method of Pathak et al. \cite{pathak2015constrained} is the only prior work that addresses deep CNNs with linear constraints in weakly supervised segmentation. It uses the constraints to synthesize fully-labeled training masks (proposals) from weak labels, mimicking full supervision and facilitating dual optimization.

        We propose to introduce a differentiable penalty, which enforces inequality constraints directly in the loss function, avoiding expensive Lagrangian dual iterates and proposal generation. From constrained-optimization perspective, our simple penalty-based approach is not optimal as there is no guarantee that the constraints are satisfied. However, surprisingly, it yields {\em substantially} better results than the Lagrangian-based constrained CNNs in \cite{pathak2015constrained}, while reducing the computational demand for training.
        \rred{By annotating only a small fraction of the pixels, the proposed approach can reach a level of segmentation performance that is comparable to full supervision on three separate tasks.}
        %\rred{Using only a fraction of the annotated pixels, we reached segmentation performances close to full supervision on three separate tasks.}
        % }In the context of cardiac images, we reached a segmentation performance close to full supervision using a fraction  of the full ground-truth labels ($0.1\%$).
        %($0.1\%$ of the pixels of the ground-truth masks) and image-level tags}.
        While our experiments focused on basic linear constraints such as the target-region size and image tags, our framework can be easily extended to other non-linear constraints, e.g., invariant shape moments \cite{Klodt2011} and other region statistics \cite{Lim2014}. Therefore, it has the potential to close the gap between weakly and fully supervised learning in semantic medical image segmentation. Our code is publicly available.
    \end{abstract}

    % \begin{keyword}
    %     Deep learning, semantic segmentation, weakly-supervised learning, CNN constraints.
    % \end{keyword}

    % \end{frontmatter}

    % \linenumbers

%--------------------------------
    \section{Introduction}
        \label{intro}

        In the recent years, deep convolutional neural networks (CNNs) have been dominating semantic segmentation problems, both in computer vision  and medical imaging, achieving ground-breaking performances when full-supervision is available \cite{FCN,dolz20173d,Litjens2017}. In semantic segmentation, full supervision requires laborious pixel/voxel annotations, which may not be available in a breadth of applications, more so when dealing with volumetric data. \textcolor{black}{Furthermore, pixel/voxel level annotations become a serious impediment for scaling deep segmentation networks to new object categories or target domains.}

        \textcolor{black}{To reduce the burden of pixel-level annotations,} weak supervision in the form partial or uncertain labels, for instance, bounding boxes \cite{dai2015boxsup}, points \cite{Bearman2016}, scribbles \cite{scribblesup,ncloss:cvpr18}, or image tags \cite{pinheiro2015image,wei2017stc}, is attracting significant research attention.
        Imposing prior knowledge on the network’s output in the form of unsupervised loss terms is a well-established approach in machine learning \cite{weston2012deep,goodfellow2016deep}. Such priors can be viewed as regularization terms that leverage unlabeled data, embedding domain-specific knowledge. For instance, the recent studies in \cite{tang2018regularized,ncloss:cvpr18} showed that direct regularization losses, e.g., dense conditional random field (CRF) or pairwise clustering, can yield outstanding results in weakly supervised segmentation, reaching almost full-supervision performances in natural image segmentation. Surprisingly, such a principled direct-loss approach is not common in weakly supervised segmentation. In fact, most of the existing techniques synthesize fully-labeled training masks (proposals) from the available partial labels, mimicking full supervision \cite{deepcut,papandreou2015weakly,scribblesup,kolesnikov2016seed}. Typically, such proposal-based techniques iterate two steps: CNN learning and proposal generation facilitated by dense CRFs and fast mean-field inference \cite{koltun:NIPS11}, which are now the de-facto choice for pairwise regularization in semantic segmentation algorithms.

        Our purpose here is to embed high-order (global) inequality constraints on the network outputs directly in the loss function, so as to guide learning. For instance, assume that we have some prior knowledge on the size (or volume) of the target region, e.g., in the form of lower and upper bounds on size, a common scenario in medical image segmentation \cite{Niethammer2013,Gorelick2013}. Let $I:\Omega  \subset \mathbb{R}^{2,3} \rightarrow \mathbb{R}$ denotes a given training image, with $\Omega$ a discrete image domain and $|\Omega|$ the number of pixels/voxels in the image.
        $\Omega_L \subseteq \Omega$ is a weak (partial) ground-truth segmentation of the image, taking the form of a partial annotation of the target region, e.g., a few points (see Figure \ref{fig:weakLab}).
        %or image-level tags.
        In this case, one can optimize a {\em partial} cross-entropy loss subject to inequality constraints on the network outputs \cite{pathak2015constrained}:
        \begin{equation}
        \label{Constrained-CNN}
        \min_{\ttt} \; \mathcal{H}(S)  \quad \mbox{s.t} \quad a \leq \sum_{p \in \Omega} S_p   \leq b
        \end{equation}
        where S = $(S_1, \dots, S_{|\Omega|}) \in [0,1]^{|\Omega|} $ is a vector of softmax probabilities\footnote{The softmax probabilities take the form: $S_p(\ttt, I) \propto \exp f_p(\ttt, I)$, where $f_p(\ttt,I)$ is a real scalar function representing the output of the network for pixel $p$. For notation simplicity, we omit the dependence of $S_p$ on $\ttt$ and $I$ as this does not result in any ambiguity in the presentation.} generated by the network at each pixel $p$ and $\mathcal{H}(S) = -\sum_{p \in \Omega_L} \log(S_p)$.
        %   , where $Y_p$ denotes one-hot encoding of ground-truth of the labeled pixels.
        Priors $a$ and $b$ denote the given upper and lower bounds on the size (or cardinality) of the target region. Inequality constraints of the form in \eqref{Constrained-CNN} are very flexible because they do not assume exact knowledge of the target size, unlike \cite{Zhang2017,Boykov2015,jia2017constrained}.
        Also, multiple instance learning (MIL) constraints \cite{pathak2015constrained}, which enforce image-tag priors, can be handled by constrained model \eqref{Constrained-CNN}. Image tags are a form of weak supervision, which enforce the constraints that a target region is present or absent in a given training image \cite{pathak2015constrained}. They can be viewed as particular cases of the inequality constraints in \eqref{Constrained-CNN}. For instance, a suppression constraint, which takes the form $\sum_{p \in \Omega} S_p \leq 0$, enforces that the target region is not in the image. $\sum_{p \in \Omega} S_p \geq 1$ enforces the presence of the region.

        Even though constraints of the form \eqref{Constrained-CNN} are linear (and hence convex) with respect to the network outputs, constrained problem \eqref{Constrained-CNN} is very challenging due to the non-convexity of CNNs. One possibility would be to minimize the corresponding Lagrangian dual. However, as
        pointed out in \cite{pathak2015constrained, Marquez-Neila2017}, this is computationally intractable for semantic segmentation networks involving millions of parameters; one has to optimize a CNN within each dual iteration. In fact, constrained optimization has been largely avoided in deep networks \cite{Ravi2018}, even thought some Lagrangian techniques were applied to neural networks a long time before the deep learning era \cite{Zhang1992,Platt1988}. These constrained optimization techniques are not applicable to deep CNNs as they solve large linear systems of equations. The numerical solvers underlying these constrained techniques would have to deal with matrices of very large dimensions in the case of deep networks \cite{Marquez-Neila2017}.

        To the best of our knowledge, the method of Pathak et al. \cite{pathak2015constrained} is the only prior work that addresses inequality constraints in deep weakly supervised CNN segmentation. It uses the constraints to synthesize fully-labeled training masks (proposals) from the available partial labels, mimicking full supervision, which avoids
        intractable dual optimization of the constraints when minimizing the loss function. The main idea of \cite{pathak2015constrained} is to model the proposals via a latent distribution. Then, it minimize a KL divergence, encouraging the softmax output of the CNN to match the latent distribution as closely as possible.
        Therefore, they impose constraints on the latent distribution rather than on the network output, which facilitates Lagrangian dual optimization. This decouples stochastic gradient descent learning of the network parameters and constrained optimization: The authors of \cite{pathak2015constrained} alternate between optimizing w.r.t the latent distribution, which corresponds to proposal generation subject to the constraints\footnote{This sub-problem is convex when the constraints are convex.}, and standard stochastic gradient descent for optimizing w.r.t the network parameters.

        We propose to introduce a differentiable term, which enforces inequality constraints \eqref{Constrained-CNN} directly in the loss function, avoiding expensive Lagrangian dual iterates and proposal generation. From constrained optimization perspective, our simple approach is not optimal as there is no guarantee that the constraints are satisfied.
        However, surprisingly, it yields {\em substantially} better results than the Lagrangian-based constrained CNNs in \cite{pathak2015constrained}, while reducing the computational demand for training. In the context of cardiac image segmentation, we reached a performance close to full supervision while using a fraction of the full ground-truth labels ($0.1\%$).
        %($0.1\%$ of the pixels of the ground-truth masks) with image-level tags}.
        Our framework can be easily extended to non-linear inequality constraints, e.g., invariant shape moments \cite{Klodt2011} or other region statistics \cite{Lim2014}. Therefore, it has the potential to close the gap between weakly and fully supervised learning in semantic medical image segmentation. Our code is publicly available \footnote{The code can be found at \url{https://github.com/LIVIAETS/SizeLoss_WSS}}.

%--------------------------------
    \bblue{\section{Related work}}

    \subsection{Weak supervision for semantic image segmentation:}
        %\paragraph{\textbf{Weak supervision for semantic image segmentation}}
        \bblue{Training segmentation models with partial and/or uncertain annotations is a challenging problem \cite{vezhnevets2011weakly,buhmann2012weakly}. Due to the relatively easy task of providing global, image-level information about the presence or absence of objects in an image, many weakly supervised approaches used image tags to learn a segmentation model \cite{verbeek2007region,vezhnevets2010towards}. For example, in \cite{verbeek2007region}, a probabilistic latent semantic analysis (PLSA) model was learned from image-level keywords. This model was later employed as a unary potential in a Markov random field (MRF) to capture the spatial 2D relationships between neighbours. Also, bounding boxes have become very popular as weak annotations due, in part, to the wide use of classical interactive segmentation approaches such as the very popular GrabCut \cite{rother2004grabcut}. This method learns two Gaussian mixture models (GMM) to model the foreground and background regions defined by the bounding box. To segment the image, appearance and smoothness are encoded in a binary MRF, for which exact inference via graph-cuts is possible, as the energies are sub-modular. Another popular form of weak supervision is the use of scribbles, which might be performed interactively by an annotator so as to correct the segmentation outcome.}
        %\bigskip

        \bblue{GrabCut is a notable example in a wide body of ``shallow'' interactive segmentation works  that used weak supervision before the deep learning era. More recently, within the computer vision community, there has been a substantial interest in leveraging weak annotations to train deep CNNs for color image segmentation using, for instance, image tags \cite{pathak2015constrained, pathak2014fully,xu2014tell,papandreou2015weakly,pinheiro2015image,wei2017stc}, bounding boxes \cite{dai2015boxsup,deepcut,khoreva2017simple}, scribbles \cite{xu2015learning,scribblesup,vernaza2017learning,tang2018regularized,ncloss:cvpr18} or points \cite{Bearman2016}.
        Most of these weakly supervised semantic segmentation techniques mimic full supervision by generating full training masks (segmentation proposals) from the weak labels. The proposals can be viewed as synthesized ground-truth used to train a CNN. In general, these techniques follow an iterative process that alternates two steps: (1) standard stochastic gradient descent for training a CNN from the proposals; and (2) standard regularization-based segmentation, which yields the proposals. This second step typically uses a standard optimizer such mean-field inference \cite{papandreou2015weakly,deepcut} or graph cuts \cite{scribblesup}. In particular, the dense CRF regularizer of Kr{\"{a}}henb{\"{u}}hl and Koltun \cite{koltun:NIPS11}, facilitated by fast parallel mean-field inference, has become very popular in semantic segmentation, both in the fully \cite{Arnab2018,Chen2015a} and weakly \cite{papandreou2015weakly,deepcut} supervised settings. This followed from the great success of DeepLab \cite{Chen2015a}, which popularized the use of dense CRF and mean-field inference as a post-processing step in the context  fully supervised CNN segmentation.}
        %\bigskip

        \bblue{An important drawback of these proposal strategies is that they are vulnerable to errors in the proposals, which might reinforce themselves in such self-taught learning schemes \cite{chapelle2006semi}, undermining convergence guarantee. The recent approaches in \cite{tang2018regularized,ncloss:cvpr18} have integrated standard regularizers such as dense CRF or pairwise graph clustering directly into the loss functions, avoiding extra inference steps or proposal generation. Such direct regularization losses achieved state-of-the-art performances for weakly supervised color segmentation, reaching near full-supervision accuracy. While these approaches encourage pairwise consistencies between pixels during training, they do not explicitly impose global constraint as in \eqref{Constrained-CNN}.}

    \subsection{Medical image segmentation with weak supervision:}
        %\paragraph{\textbf{Medical image segmentation with weak supervision}}
        \bblue{Despite the increasing amount of works focusing on weakly supervised deep CNNs in semantic segmentation of color images, leveraging weak annotations in medical imaging settings is not simple. To our knowledge, the literature on this matter is still scarce, which makes weak-supervision approaches appealing in medical image segmentation. As in color images, common settings for weak annotations are bounding boxes. For instance, DeepCut \cite{deepcut} follows a similar setting as \cite{papandreou2015weakly}. It generates image proposals, which are refined by a dense CRF before being re-used as ``fake'' labels to train the CNN.
        Using the bounding boxes as initializations for the Grab-cut algorithm, the authors showed that, by this iterative optimization scheme, one can obtain a performance better than the shallow counterpart, i.e., GrabCut. In another weakly supervised scenario \cite{rajchl2016learning}, images were segmented in an unsupervised manner, generating a set of super-pixels \cite{achanta2012slic}, among which users had to select the regions belonging to the object of interest.
        Then, these masks generated from the super-pixels were employed to train a CNN. Nevertheless, as proposals are generated in an unsupervised manner, and due to the poor contrast and challenging targets typically present in medical images, these ``fake'' labels are likely prone to errors, which can be propagated during training, as stated before.}

    \subsection{Constrained CNNs:}
        %\paragraph{\textbf{Constrained CNNs}}
        \bblue{To the best of our knowledge, there are only a few recent works \cite{pathak2015constrained, Marquez-Neila2017, jia2017constrained} that addressed imposing global constraints on deep CNNs.
        In fact, standard Lagrangian-dual optimization has been completely avoided in modern deep networks involving millions of parameters. As pointed out recently in \cite{pathak2015constrained, Marquez-Neila2017}, there
        is a consensus within the community that imposing constraints on the outputs of deep CNNs that are common in modern computer vision and medical image analysis problems is impractical: the direct use of Lagrangian-dual optimization for networks with millions of parameters requires training a whole CNN after each iterative dual step \cite{pathak2015constrained}. To avoid computationally intractable dual optimization, Pathak et al. \cite{pathak2015constrained} imposed inequality constraints on a latent distribution instead of the network output. This latent distribution describes a ``fake'' ground truth (or segmentation proposal). Then, they trained a single CNN so as to minimize the KL divergence between the network probability outputs and the latent distribution. This prior-art work is the most closely related to our study and, to our knowledge, is the only work that addressed inequality constraints in weakly supervised CNN segmentation. The work in \cite{Marquez-Neila2017} imposed hard equality constraints on 3D human pose estimation. To tackle the computational difficulty, they used a Kyrlov sub-space approach and limited the solver to only a randomly selected sub-set of the constraints within each iteration. Therefore, constraints that are satisfied at one iteration may not be satisfied at the next, which might explain the negative results in \cite{Marquez-Neila2017}. A surprising result in \cite{Marquez-Neila2017} is that replacing the equality constraints with simple $L_2$ penalties yields better results than Lagrangian optimization, although such a simple penalty-based formulation does not guarantee constraint satisfaction. A similar $L_2$ penalty was used in \cite{jia2017constrained} to impose equality constraints on the size of the target regions in the context of histopathology segmentation. While the equality-constrained formulations in \cite{Marquez-Neila2017, jia2017constrained} are very interesting, they assume exact knowledge of the target function (e.g., region size), unlike the inequality-constraint formulation in \eqref{Constrained-CNN}, which allows much more flexibility as to the required prior domain-specific knowledge.}

%--------------------------------------
    \section{Proposed loss function}
        \label{sec:methods}

\begin{figure}[h!]
            \centering
            \includegraphics[width=0.4\linewidth]{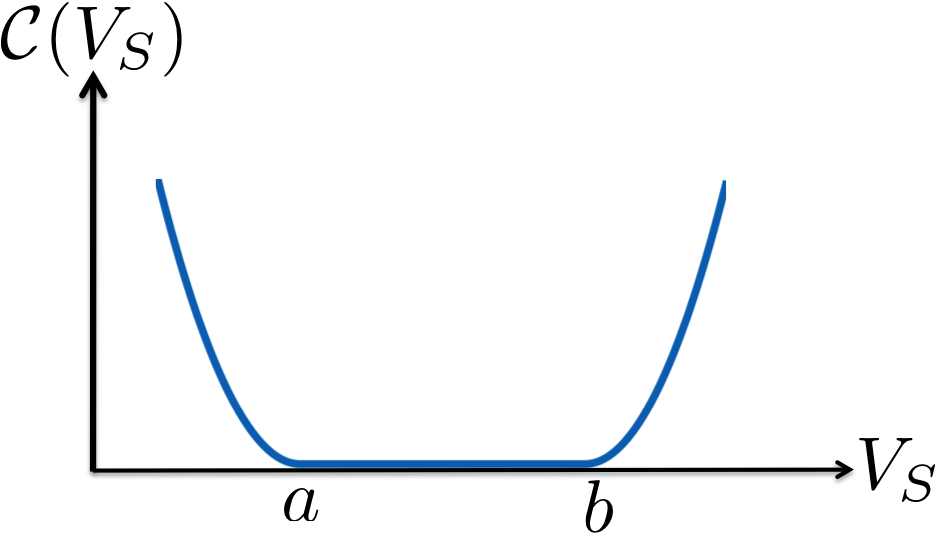}
            \caption{Illustration of our differentiable loss for imposing soft size constraints on the target region.}
            \label{fig:sizeConstraint}
        \end{figure}

        We propose the following loss for weakly supervised segmentation:
        \begin{equation}
            \label{Cross-Entropy}
            % - \sum_{p \in \Omega_L} Y_p \log(S_p)
            \mathcal{H}(S)
            + \lambda \, {\cal C} \, (V_S),
        \end{equation}
        where $V_S = \sum_{p \in \Omega} S_p$, $\lambda$ is a positive constant that weighs the importance of constraints, and function ${\cal C}$ is given by (See the illustration in Fig. \ref{fig:sizeConstraint}):
        \begin{equation}
            \label{eq:lossForward1}
            \mathcal{C}(V_S) =
            \begin{cases}
                \left( V_S - a \right)^2, & \text{if}\ V_S < a\\
                \left( V_S - b \right)^2, & \text{if}\ V_S > b\\
                0, & \text{otherwise} \\
            \end{cases}
        \end{equation}
        Now, our differentiable term ${\cal C}$ accommodates standard stochastic gradient descent.
        During back-propagation, the term of gradient-descent update corresponding to ${\cal C}$ can
        be written as follows:
        \begin{equation}
            \label{eq:lossBackward1}
            - \frac{\partial \mathcal{C}(V_S)}{\partial \ttt}  \propto
            \begin{cases}
                \left (a - V_S \right) \frac{\partial S_p}{\partial \ttt}, & \text{if}\ V_S < a\\
                \left (b - V_S \right) \frac{\partial S_p}{\partial \ttt}, & \text{if}\ V_S > b\\
                0, & \text{otherwise} \\
            \end{cases}
        \end{equation}
        where $\frac{\partial S_p}{\partial \ttt}$ denotes the standard derivative of the softmax outputs of the network. The gradient in \eqref{eq:lossBackward1}
        has a clear interpretation. During back-propagation, when the current constraints are satisfied, i.e., $a \leq V_S \leq b$, observe that
        $\frac{\partial \mathcal{C}(V_S)}{\partial \ttt} = 0$. Therefore, in this case, the gradient stemming from our term has no effect on the current update of the network parameters. Now, suppose without loss of generality that the current set of parameters $\ttt$ corresponds to $V_S < a$, which means the current target region is smaller than its lower bound $a$. In this case of constraint violation, term $(a - V_S)$ is positive and, therefore, the first line of \eqref{eq:lossBackward1} performs a gradient {\em ascent} step on softmax outputs, increasing $S_p$. This makes sense because it increases the size of the current region, $V_S$, so as to satisfy the constraint. The case $V_S > b$ has a similar interpretation.

        The next section details the dataset, the weak annotations and our implementation. Then, we report comprehensive evaluations of the effect of our constrained-CNN losses on segmentation performance. We also report comparisons to the Lagrangian-based constrained CNN method in \cite{pathak2015constrained} and to the fully supervised setting.

%--------------------------------------
    \section{Experiments}

        \subsection{Medical Image Data:}
        \rred{In this section, the proposed loss function is evaluated on three publicly available datasets, each corresponding to a different application -- cardiac, vertebral body and prostate segmentation. Below are additional  details of these data sets.}

            \subsubsection{Left-ventricle (LV) on cine MRI}
                A part of our experiments focused on left ventricular endocardium segmentation. We used the training set from the publicly available data of the 2017 ACDC Challenge\footnote{\url{https://www.creatis.insa-lyon.fr/Challenge/acdc/}}. This set consists of 100 cine magnetic resonance (MR) exams covering well defined pathologies: dilated cardiomyopathy, hypertrophic cardiomyopathy, myocardial infarction with altered left ventricular ejection fraction and abnormal right ventricle. It also included normal subjects. \rred{Each exam contains acquisitions only at the diastolic and systolic phases.} The exams were acquired in breath-hold with a retrospective or prospective gating and a SSFP sequence in 2-chambers, 4-chambers and in short-axis orientations. A series of short-axis slices cover the LV from the base to the apex, with a thickness of 5 to 8 mm and an inter-slice gap of 5 mm. The spatial resolution goes from 0.83 to 1.75 mm$^2$/pixel. For all the experiments, we employed the same 75 exams for training and the remaining 25 for validation.

                %\rred{Two images (systole and diastole) were available per patient, giving a final 150/50 images for training/validation.}

                %\rred{Since each patient is two volumes, this gives us 150/50 for training/validation.}

            \subsubsection{Vertebral body (VB) on MR-T2}
                This dataset contains 23 3D T2-weighted turbo spin echo MR images from 23 patients and the associated ground-truth segmentation, and is freely available from \footnote{\url{http://dx.doi.org/10.5281/zenodo.22304}}. Each patient was scanned with 1.5 Tesla MRI Siemens scanner (Siemens Healthcare, Erlangen, Germany) to generate T2-weighted sagittal images. All the images are sampled to have the same sizes of 39$\times$305$\times$305 voxels, with a voxel spacing of 2$\times$1.25$\times$1.25 mm$^3$. In each image, 7 vertebral bodies, from T11 to L5, were manually identified and segmented, resulting in 161 labeled regions in total. For this dataset, we employed 15 scans for training and the remaining 5 for validation.
                %%%%% JOSE: Maybe the second clinical information is not needed %%%%

            \rred{\subsubsection{Prostate segmentation on MR-T2}}
                \rred{The third dataset was made available at the MICCAI 2012 prostate MR segmentation challenge\footnote{\url{https://promise12.grand-challenge.org}}. It contains the transversal T2-weighted MR images of 50 patients acquired at different centers with multiple MRI vendors and different scanning protocols. It is comprised of various diseases, i.e., benign and prostate cancers.
                The images resolution ranges from $15\times256\times256$ to $54\times512\times512$ voxels with a spacing ranging from $2\times0.27\times0.27$ to $4\times0.75\times0.75$mm$^3$. We employed 40 patients for training and 10 for validation.}

        \subsection{Weak annotations:}
        \label{sssec:weaklyLabels}

\begin{figure}[h!]
                \includegraphics[width=1\textwidth]{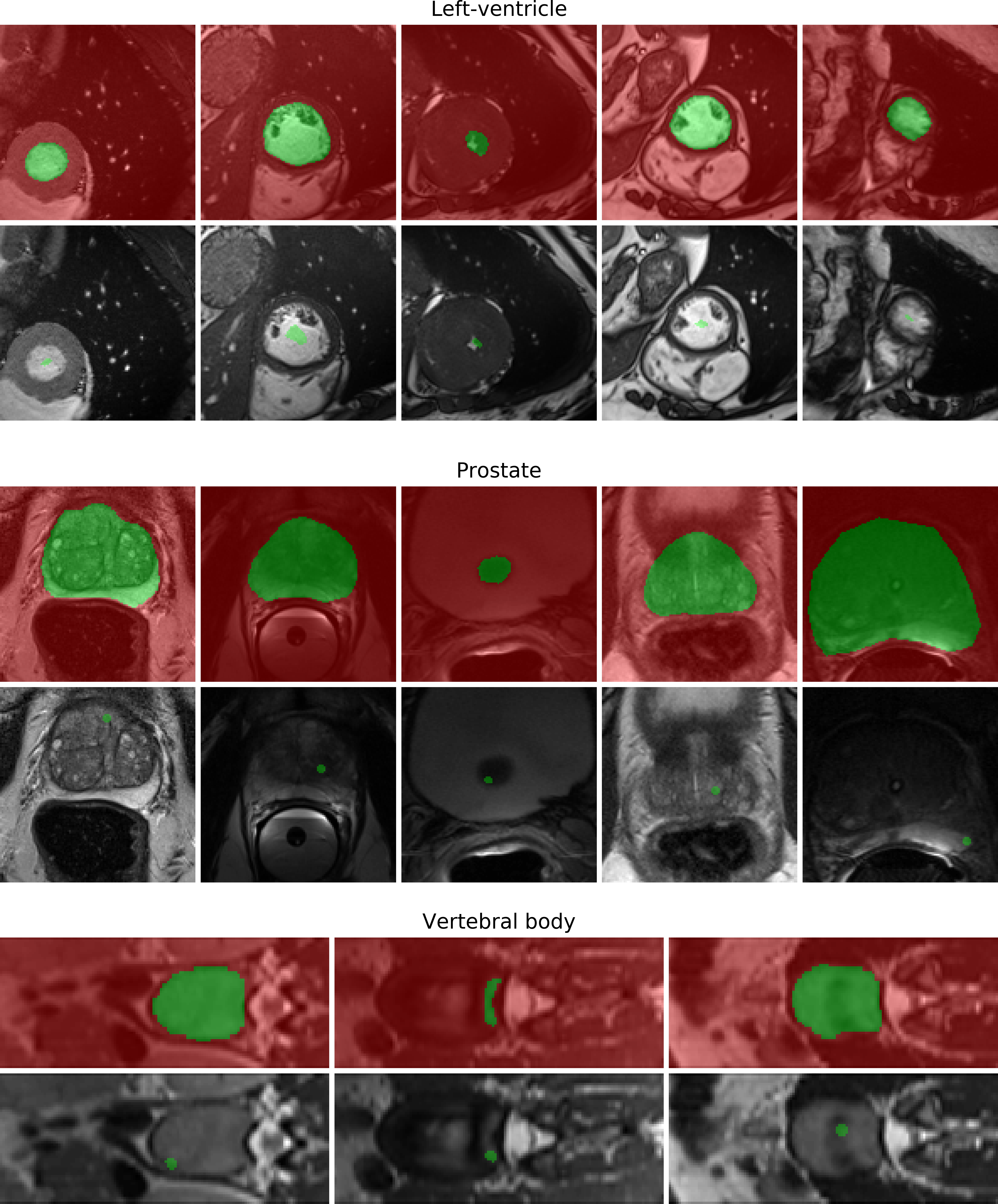}
                \caption{Examples of different levels of supervision. In the fully labeled images (\textit{top}), all pixels are annotated, with red depicting the background and green the region of interest. In the weakly supervised cases (\textit{bottom}), only the labels of the green pixels are known. The images were cropped for a better visualization of the weak labels. The original images are of size 256 $\times$ 256 pixels.}
                \label{fig:weakLab}
            \end{figure}

            \rred{To show that the proposed approach is robust to the strategy for generating the weak labels, as well as to their location, we consider two different strategies generating weak annotations from fully labeled images. Figure \ref{fig:weakLab} depicts some examples of fully annotated images and the corresponding weak labels.}

            \paragraph{Erosion} \bblue{For the left-ventricle dataset}, we employed binary erosion on the fully annotations with a kernel of size 10$\times$10. If the resulted label disappeared, we repeated the operation with a smaller kernel (i.e., 7$\times$7) until we get a small contour. Thus, the total number of annotated pixels represented the 0.1$\%$ of the labeled pixels in the fully supervised scenario. This correspond to the second row in Figure \ref{fig:weakLab}.

            % To show the robustness of the proposed approach to seeds placement, t
            \paragraph{Random point} \rred{The weak labels for the vertebral body and prostate datasets were generated by randomly selecting a point within the ground-truth mask and creating a circle around it with a maximum radius of 4 pixels (fourth and sixth row in Fig. \ref{fig:weakLab}), while ensuring there is no overlap with the background. With these weak annotations, only $0.02\%$ of the pixels in the dataset have ground-truth labels.}

            % \begin{figure}[h]
            %     \begin{subfigure}[b]{1\textwidth}
            %         \includegraphics[width=1\textwidth]{Images/prostate_labels}
            %         \caption{\rred{Generated labels for the prostate dataset.}}
            %     \end{subfigure}
            %     \begin{subfigure}[b]{1\textwidth}
            %         \includegraphics[width=1\textwidth]{Images/spine_labels}
            %         \caption{\rred{Generated labels for the spine dataset.}}
            %     \end{subfigure}
            %     % \caption{\rred{Comparison between the full masks of the Spine dataset, and the weak labels that we generated. The strategy was to put points randomly in the target area, in order to mimic what a human annotator could do in a weakly supervised setting}}
            %     \caption{\rred{Comparison between the full mask and the weak labels created, when using random points as generation strategy. Best viewed in colors.}}
            %     \label{fig:prostate_spine_labels}
            % \end{figure}

        \subsection{Different levels of supervision:}
            \rred{Training models with diverse levels of supervision requires that appropriate objectives be defined for each case. In this section, we introduce the different models, each with different levels of supervision.}
            %We first define the baselines we will compare to, before defining the bounds we will be using for the constrained optimization. We then define the protocol of different ablation studies we performed.}

            \subsubsection{Baselines}
                \rred{We trained a segmentation network from weakly annotated images with no additional information, which served as a lower baseline. Training this model relies on minimizing the cross-entropy corresponding to the fraction of labeled pixels: $\mathcal{H}(S) = -\sum_{p \in \Omega_L} \log(S_p)$. In the following discussion of the experiments, we refer to this model as \textit{partial cross-entropy (CE)}.}

                \rred{As an upper baseline, we resort to the fully-supervised setting, where class labels (foreground and background) are known for every pixel during training ($\Omega_L = \Omega$). This model is referred to as \textit{fully-supervised}.}

            \subsubsection{Size constraints}
                \label{sec:bounds_values}
                \rred{We incorporated information about the size of the target region during training, and optimized the partial cross-entropy loss subject to inequality constraints of the general form in Eq. \eqref{Constrained-CNN}.
                We trained several models using the same weakly annotated images but different constraint values.}

                % The same bounds are used for our method and the Lagrangian proposal in \cite{pathak2015constrained}.

                \paragraph{Image tags bounds}
                    \rred{Similar to MIL scenarios, we first used image-tag priors by enforcing the presence or absence of a the target in a given training image, as introduced earlier. This reduces to enforcing that the size of the predicted region is less or equal to $0$ if the target is absent from the image, or larger than $0$ otherwise. To simplify the implementation, we can represent the constraints as:}
                    \begin{equation}
                        \label{eq:tag_bounds}
                         a, b = \begin{cases} 1, |\Omega| &\text{if target is present } (\Omega_L \neq \emptyset) \\ 0, 0 &\text{otherwise}\end{cases} .
                    \end{equation}

                    \rred{While being very coarse, these constraints convey relevant information about the target regions, which
                    may be used to find common patterns in the case of region absence or presence.}

                    % , already have good results while requiring no additionnal work or expert knowledge, as we can deduce them from the weak labels.

                \paragraph{Common bounds}
                    \rred{The next level of supervision consists of using tighter bounds for the positive cases, instead of $(1, |\Omega|)$. To this end, the complete segmentation of a {\em single} patient is employed to compute the minimum and maximum size of the target region across all the slices. Then, we multiplied these minimum and maximum values by $0.9$ and $1.1$, respectively, to account for inter-patient variability. In this case, all the images containing the object of interest have the same lower and upper bounds. As an example, this results in the following values for the ACDC dataset:}
                    \begin{equation}
                        \label{eq:common_bounds}
                         a, b = \begin{cases} 60, 2000 &\text{if target is present } (\Omega_L \neq \emptyset) \\ 0, 0 &\text{otherwise}\end{cases} .
                    \end{equation}
                    %\rred{This means that all slices for all patients have the same bounds. Depending on the task, the range of values can be fairly wide, and there not provide much more information that tags. On the contrary, very structured problems without much variation between slices can have much tighter bounds.}

                \paragraph{Individual bounds}
                    \rred{
                    %A downside of employing
                    With common bounds, the range of values for a given target may be very large.
                    %To overcome this issue
                    To investigate whether a more precise knowledge of the target is helpful, we also consider the use of individual bounds for each slice, based on the true size of the region:
                    \[\tau_Y = \sum_{p \in \Omega} Y_p, \]
                    with $Y = (Y_1, ..., Y_{|\Omega|}) \in \{0, 1\}^{|\Omega|}$ denoting the full annotation of image $I$. As before, we introduce some uncertainty on the target size, and multiply $\tau_Y$ by the same lower and upper factors, resulting in the following bounds:}
                    \begin{equation}
                        \label{eq:indiv_bounds}
                         a, b = \begin{cases} 0.9\tau_Y, 1.1\tau_Y &\text{if target is present } (\Omega_L \neq \emptyset) \\ 0, 0 &\text{otherwise}\end{cases} .
                    \end{equation}
                    %\begin{equation}
                    %    a, b = 0.9\tau_Y, 1.1\tau_Y .
                    %\end{equation}
                    %\rred{Since we use the true size of the object as a base to compute our bounds, we do not need to handle the case for positive and negatives slices explicitly ($\tau=0$ when there is no object). }

             \subsubsection{Hybrid training}
                \rred{We also investigate whether combining our proposed weak supervision approach with fully annotated images during the training leads to performance improvements. For this purpose, considering we have a training set of $m$ weakly annotated images, we replace $n$ ($n<m$) among these by their fully annotated counterparts. Thus, the training amounts to minimizing the cross-entropy loss for the $n$ fully annotated images, along with the partial cross-entropy constrained with common size bounds for the remaining $m-n$ weakly labeled images. To examine the positive effect of size constraints in this scenario (referred to as \textit{Hybrid}), we compare the results to a network trained with the $n$ fully annotated images (without constraints).}

                % In theory, this should improve the performances of the trained model, reducing the gap with full-supervision.

            \subsection{Constraining a 3D volume:}
               % \rred{
            %    While in theory it is trivial to extend the proposed formulation for 3D images, memory limitations of current hardware make this task difficult. Methods sampling sub-patches to perform 3D-convolution, such as \cite{dolz20173d,dolz2018hyperdense}, cannot be used here, as prediction of a sub-patch is not representative of the whole targeted volume and processing all the sub-patches of a volume in a single pass is not possible with our available resources.}
            %    \rred{To circumvent this limitation, we decided to perform segmentation on independent 2D slices, with all the slices of the 3D image contained in the same batch. Thus, we are able to compute a 3D volume while staying within our memory budget.}

                \rred{We can extend our formulation to constrain a 3D volume as follows:
                \begin{align*}
                    \sum_{S \in \text{B}} & \mathcal{H}(S) + \lambda \mathcal{C}(V_\text{B}), \qquad \textrm{with} \quad V_{\text{B}} = \sum_{S \in \text{B}} V_S \\
                        %& \tau_\text{B} = \sum_{Y \in \text{B}} \tau_Y \\
                        %& a, b = 0.9\tau_\text{B}, 1.1\tau_\text{B}
                \end{align*}
                where $V_{\text{B}}$ denotes the target-region volume, $\text{B} = ((Y^1, S^1), ..., (Y^{|B|}, S^{|B|}))$ denotes a training batch containing all the 2D slices of the 3D volume\footnote{For readability, we simplify a batch as a list of labels $Y$ and associated predictions $S$.}, and the 3D constraints are now given by:}
                \begin{align*}
                        & a, b = 0.9\tau_\text{B}, 1.1\tau_\text{B}, \qquad
                         \textrm{with} \quad \tau_\text{B} = \sum_{Y \in \text{B}} \tau_Y \\
                \end{align*}
                \rred{Notice that, with constraints on the whole 3D volume, we have less supervision than the 2D scenarios from \ref{sec:bounds_values}, where all the 2D slices have independent supervision (e.g., the image tags).}% As we will show in the results section, this is particularly important since weak supervision in the form of the tag bounds in Eq. \eqref{eq:tag_bounds} is very helpful.} %Experiments on 3D were only performed on the ACDC dataset, as the other datasets contained too many slices per patient to be loaded in the memory at once.}

        \subsection{Training and implementation details:}
            \rred{For the experiments on the left-ventricle and vertebral-body datasets, we used ENet \cite{paszke2016enet}, as it has shown a good trade-off between accuracy and inference time. Due to the higher difficulty of the prostate segmentation task, we employed a fully residual version of U-Net \cite{ronneberger2015u}, similar to \cite{quan2016fusionnet}.}

            \rred{For the three datasets, we trained the networks from scratch using the Adam optimizer and an initial learning rate of $5 \times 10^{-4}$ that we decreased by a factor of 2 if the performances on the validation set did not improve over 20 epochs. All the 3D volumes were sliced into $256\times256$ pixels images, and zero-padded when needed. Batch sizes were equal to 1, 4, and 20 for the left-ventricle, prostate and vertebral body, respectively. Those values were not tuned for optimal performances, but to speed-up experiments when enough data were available.} The weight of our loss in \eqref{Cross-Entropy} was empirically set to 1$\times$10$^{-2}$. \rred{Due to the difficulty of the task, data augmentation was used for the prostate dataset, where we generated 4 copies of each training image using random mirroring, flipping and rotation.}

            \rred{All our tests were implemented in Pytorch \cite{paszke2017autopmatic}}. We ran the experiments on a machine equipped with a NVIDIA GTX 1080 Ti GPU (11GBs of video memory)\rred{, AMD Ryzen 1700X CPU and 32GBs of memory.} The code is available at \url{https://github.com/LIVIAETS/SizeLoss_WSS}. We used the common Dice similarity coefficient (DSC) to evaluate the segmentation performance of trained models.

                \subsubsection{\rred{Modification and tweaks for Lagrangian proposals}}
                    \rred{For a fair comparison, we re-implemented the Lagrangian-proposal method of Pathak et al. \cite{pathak2015constrained} in PyTorch, to take advantage of GPU capabilities and avoid costly transfers between GPU and CPU. Lagrangian proposals reuse the same network and loss function as the fully-supervised setting. At each iteration, the method alternates between two steps. First, it synthesizes a ground truth $\tilde Y$ with projected gradient ascent (PGA) over the dual variables, with the network parameters fixed. Then, for fixed $\tilde Y$, the cross-entropy between $\tilde Y$ and $S$ is optimized as in standard fully-supervised CNN training. The learning rate used for this PGA was set experimentally to $5 \times 10^{-5}$, as sub-optimal values lead to numerical errors. We found that limiting the number of iterations for the PGA to $500$ (instead of the original $3000$) saved time without affecting the results.}

                    \rred{We also introduced an early stopping mechanism into the PGA in the case of convergence, to improve speed without impacting the results (a comparison can be found in Table \ref{table:times}). The constraints of the form $0 \leq V_S \leq 0$ required specific care, as the formulation from \cite{pathak2015constrained} is not designed to work on equalities, unlike our penalty approach, which systematically handles equality constraints when $a=b$. In this case, the bounds for \cite{pathak2015constrained} were modified to $-1 \leq V_S \leq 0$.}

%-------------------------------
    \section{Results}
        \rred{To validate the proposed approach, we first performed a series of experiments focusing on LV segmentation. In Sec. \ref{sssec:resWeak}, the impact of including size constraints is evaluated using our direct penalty. We further compare to the Lagrangian-proposal method in \cite{pathak2015constrained}, showing that our simple method yields substantial improvements over \cite{pathak2015constrained} in the same weakly supervised settings. We also provide the results for several degrees of supervision, including hybrid and fully supervised learning in Sec. \ref{sssec:fullySup}. Then, to show the wide applicability of the proposed constrained loss, results are reported for two other applications in Sec. \ref{ssec:vb}: MR-T2 vertebral body segmentation and prostate segmentation task.  We further provide qualitative results for the three applications in Sec. \ref{sssec:qual}. In Sec. \ref{ssec:ablation_bounds}, we investigate the sensitivity of the proposed loss to both the lower and upper bounds. Finally, the efficiency of different learning strategies are compared (Sec. \ref{sssec:times}), showing that our direct constrained-CNN loss does not add to the training time, unlike the Lagrangian-proposal method in \cite{pathak2015constrained}.}
        %Furthermore, in this section we also investigated the frequency with which constraints are violated in both strategies.

        \subsection{Weakly supervised segmentation with size constraints:}
            \label{sssec:resWeak}

\begin{figure}[h!]
                \centering
                \begin{minipage}{0.49\textwidth}
                    \includegraphics[width=\textwidth]{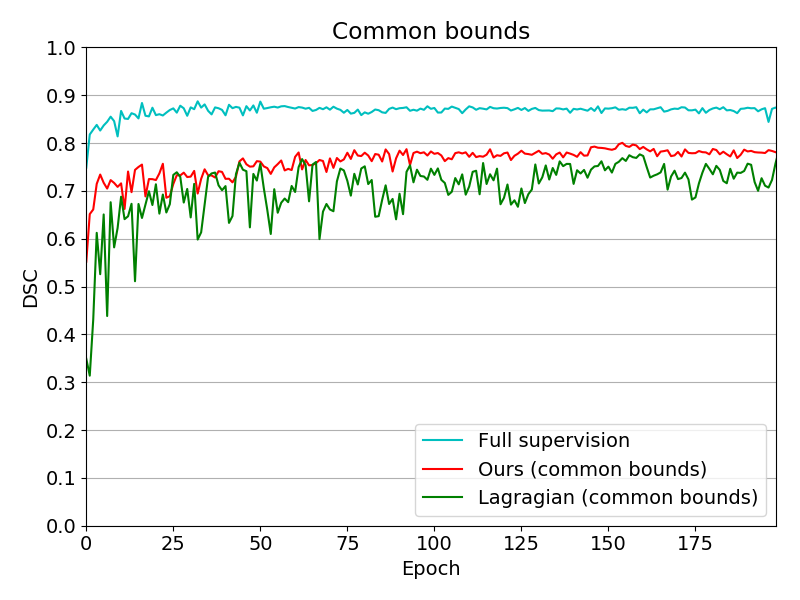}
                    % touch plot.py && make -f midl.make results/midl/val_batch_dice.png TRN="results/midl/fs results/midl/size_595 results/midl/pathak_595_norm" DEBUG="--labels 'Full supervision' 'Ours (common bounds)' 'Lagrangian (common bounds)' --ylabel DSC --title 'Common bounds' --figsize 8 6 --loc 'lower right' --fontsize 14"
                \end{minipage}
                \begin{minipage}{0.49\textwidth}
                    \includegraphics[width=\textwidth]{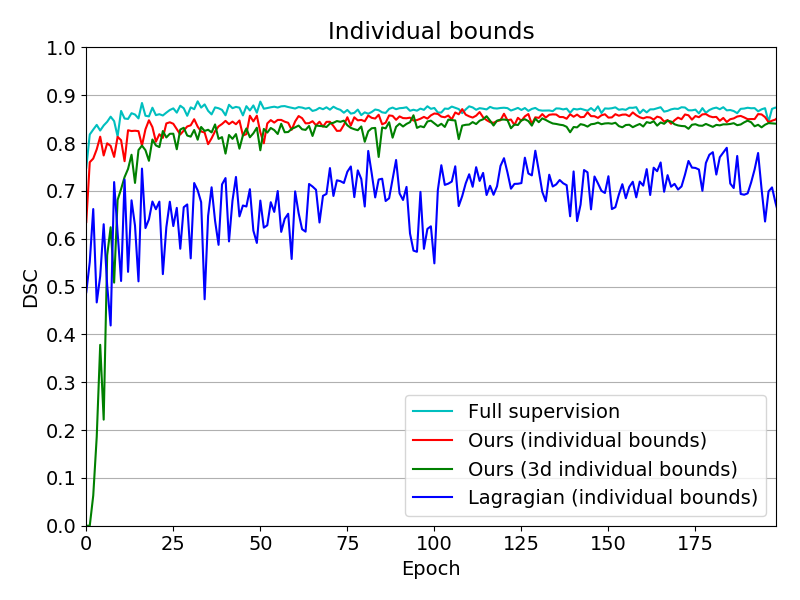}
                   % touch plot.py && make -f midl.make results/midl/val_batch_dice.png TRN="results/midl/fs results/midl/presize results/midl/3d_sizeloss results/midl/pathak_precise_norm" DEBUG="--labels 'Full supervision' 'Ours (individual bounds)' 'Ours (3d individual bounds)' 'Lagrangian (individual bounds)' --ylabel DSC --title 'Individual bounds' --figsize 8 6 --loc 'lower right' --fontsize 14"
                \end{minipage}
                \caption{\rred{Evolution of the DSC during training for the left-ventricle validation set, including the weakly supervised learning models and different strategies analyzed, with also the full-supervision setting. As tags and common bounds achieve similar results, we plot only common bounds for better readability.}}
                \label{fig:dsc_evol}
            \end{figure}

            \paragraph{\textbf{2D segmentation}} Table \ref{table:dsc_size} reports the results on the \rred{left-ventricle} validation set for all the models trained with both the Lagrangian proposals in \cite{pathak2015constrained} and our direct loss. As expected, using the partial cross entropy with a fraction of the labeled pixels yielded poor results,
            with a mean DSC less than $15\%$.
            Enforcing the image-tag constraints, as in the MIL scenarios, increased substantially the DSC to a value of $0.7924$. \rred{Using common bounds increased the results marginally in this case, slightly increasing the mean Dice value by $1\%$. The Lagrangian proposal \cite{pathak2015constrained} reaches similar results, albeit slightly lower and much more unstable than our penalty approach (see Figure \ref{fig:dsc_evol}).}

            \rred{The difference in performance is more pronounced when we employ individual bounds instead. In this setting, our method achieves a DSC of $0.8708$, only $2\%$ lower than full supervision. However, the Lagrangian-proposal method achieves a performance similar to using common (loose) bounds, suggesting that it is not able to make use of this extra, more precise information. This can be explained by its proposal-generation method, which tends to reinforce early mistakes (especially when training from scratch): the network is trained with conflicting information -- i.e., similar-looking patches are both foreground and background according the the synthetic ground truth -- and is not able to recover from those initial mis-classifications. }

            % Adding more precise constraints is not enough to offset this (Already said)

            \paragraph{\textbf{3D segmentation}}\rred{Constraining the size of the 3D volume of the target region also shows the benefit of our penalty approach, yielding a mean DSC of $0.8580$.
           % Unfortunately, we were not able to modify the method from \cite{pathak2015constrained} to test this setting.
           Recall that, here, we are using less supervision than the 2D case. Since we do not use tag information in this case, these results suggest that only a fraction of all the slices may be used when creating the labels, allowing annotators to scribble the 3D image directly instead of going through all the 2D slices one by one.}

            \begin{table}[h!]
            \centering
            \small
            \caption{Left-ventricle segmentation results with different levels of supervision. \rred{Bold font highlights the best weakly supervised setting.}}
            \begin{tabular}{cll|c}
                \multicolumn{2}{c}{Model}     & \multicolumn{1}{c}{Method} & DSC (Val) \\ \hline \hline
                \multirow{8}{*}{\begin{tabular}[c]{@{}c@{}}Weakly \\ supervised\end{tabular}} & Partial CE    & & 0.1497      \\ \cline{2-4}
                & CE + Tags & Lagrangian Proposals \cite{pathak2015constrained} & 0.7707 \\
                & Partial CE + Tags & Direct loss (Ours) & 0.7924 \\ \cline{2-4}
                & CE + Tags + Size* & Lagrangian Proposals \cite{pathak2015constrained} & 0.7854 \\
                & Partial CE + Tags + Size* & Direct loss (Ours) & 0.8004 \\ \cline{2-4}
                & CE + Tags + Size** & Lagrangian Proposals \cite{pathak2015constrained} & 0.7900 \\
                & Partial CE + Tags + Size** & Direct loss (Ours) & \textbf{0.8708} \\ \cline{2-4}
                & \rred{CE + 3D Size**} & \rred{Lagrangian Proposals \cite{pathak2015constrained}} & \rred{N/A} \\
                & \rred{Partial CE + 3D Size**} & \rred{Direct loss (Ours)} & \rred{0.8580} \\
                \hline
                \begin{tabular}[c]{@{}c@{}}Fully\\ supervised\end{tabular}    &      Cross-entropy  & &  0.8872   \\
                \bottomrule
                \multicolumn{4}{l}{\footnotesize{\rred{*Common bounds / ** Individual bounds}}}
            \end{tabular}

            \label{table:dsc_size}
            \end{table}

        \subsection{Hybrid training: mixing fully and weakly annotated images:}
            \label{sssec:fullySup}

\begin{figure}[h!]
                \centering
                \includegraphics[width=0.9\linewidth]{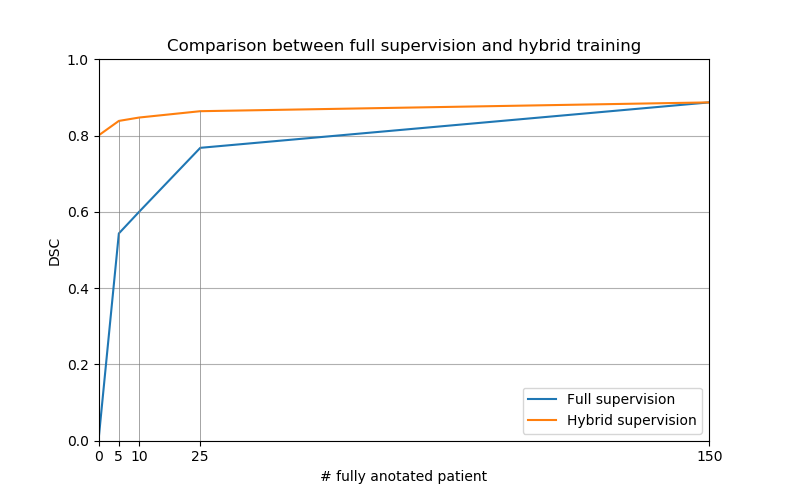}
                \caption{\rred{Mean DSC values over the number of fully annotated patients employed for training.}}
                \label{fig:hybrid}
            \end{figure}

            \rred{Table \ref{table:hybrid} and Figure \ref{fig:hybrid} summarize the results obtained when combining weak and full supervision. First, and as expected, we can observe that adding $n$ fully annotated images to the training set (Hybrid$\_n$) improves the performances in comparison to the model trained solely with the weakly annotated images, i.e., Weak$\_$All. Particularly, the DSC increases by 4\%,5\% and 6\% when $n$ is equal to 5,10 and 25, respectively, approaching the full-supervision performance with only 25\% of the fully labeled images.}

            \rred{Nevertheless, it is more interesting to see the impact of adding weakly annotated images (i.e., Hybrid$\_n$) to a model trained solely with fully labeled images (i.e., Full$\_n$). From the results, we can observe that adding weakly annotated images to the training set significantly increases the performance, particularly when the amount of fully annotated images (i.e., $n$) is limited. For instance, in the case of $n$ equal to 5, adding weakly annotated images enhanced the performance by more than 30$\%$ in comparison to full supervision with $n$ equal to $5$. Despite the fact that this gap decreases with the number of fully annotated images, the difference between both settings (i.e., Full and Hybrid) remains significant. More interestingly, training the same model with a high amount of weakly annotated images and no or a very reduced set of fully labeled images (for example Weak$\_$All or Hybrid$\_$5) achieves better performances than employing datasets with much higher numbers of fully labeled images, e.g, Full$\_$25.}

            %(+4\% DSC with 5 patients, +5\% with 10 patients and +6\% with 25 patients), almost closing the gap with full supervision ($0.8641$ compared to $0.8872$ for full supervision). The gap between Full\_$n$ and Hybrid\_$n$ tends to reduce when $n$ increases, but remains significant (almost 30\% difference with $n=5$, down to 10\% with $n=25$).}
            %\rred{Also, the performances of 150 weakly annotated images are better than 25 fully annotated images, while being much faster to produce \cite{Bearman2016}.}

            \rred{These results suggest that a good strategy when annotating a new dataset might be to start with weak labels for all the images, and progressively complete full annotations, should ressources become available.}

            \begin{table}[h!]
                \small
                \centering
                \caption{\rred{Ablation study on the amounts of fully and weakly labeled data. We report the mean DSC of all the testing cases, for all the settings and using the same architecture.}}
                \begin{tabular}{llcc}
                    \textbf{Name}            & \textbf{Training approach}     & \begin{tabular}[c]{@{}c@{}}\textbf{\# Fully/Weakly }\\  \textbf{annotated images}\end{tabular} & \textbf{DSC} \\
                    \toprule \toprule
                    Weak\_All &   Weak supervision*   & 0/150 & 0.8004 \\
                    \midrule
                    Full\_5    & Full supervision  & 5/0 &  0.5434   \\
                    Hybrid\_5  & Full + weak supervision* & 5/145 &  0.8386   \\ \midrule
                    Full\_10   & Full supervision  & 10/0 & 0.6004    \\
                    Hybrid\_10 & Full + weak supervision* & 10/140 & 0.8475    \\ \midrule
                    Full\_25   & Full supervision   & 25/0 & 0.7680 \\
                    Hybrid\_25 & Full + weak supervision* & 25/125 &  0.8641  \\ \midrule
                    Full\_All &   Full supervision   & 150/0 & 0.8872 \\  \bottomrule
                    \footnotesize{\rred{*Common bounds}}
                \end{tabular}
                \label{table:hybrid}
            \end{table}

        \subsection{MR-T2 vertebral body and prostate segmentation:}
        \label{ssec:vb}
            % Results comes from archives/constrained_cnn-190128-25947f0-newton-zenodo_spine.tar.gz
            % results comes from archives/constrained_cnn-190102-0a7f058-feynman-prostate.tar.gz

            \rred{The results obtained for the vertebral-body dataset (Table \ref{table:dices_spine_prostate}) highlight well the differences in the performances of different levels of supervision. Using tag bounds produces a network that roughly locates the object of interest (DSC of $0.5597$), but fails to identify its boundaries (as seen in Figure \ref{fig:spine_results}, \textit{third column}). Employing the common size strategy achieves satisfactory results for the slices containing objects with a regular shape but still fails when more difficult/irregular targets are present, resulting in an overall improvement of DSC ($0.7900$). However, when using individual bounds, the network is able to satisfactory segment even the most difficult cases, obtaining a DSC of $0.8604$, only $3\%$ lower than full supervision.}

            \begin{table}[h!]
                \centering
                \caption{\bblue{Mean Dice scores (DSC) for several degrees of supervision}\rred{, using the vertebral-body and prostate validation sets. Bold font indicates the best weakly
                supervised setting for each data set.}}
                \begin{tabular}{l|c|c}  \hline
                    \textbf{Method}                             & \textbf{Vertebral body DSC} & \textbf{Prostate DSC} \\
                    \hline
                    Partial CE                          & 0.1155     & 0.0320         \\
                    Partial CE + Tags                   & 0.5597     & 0.6911 \\
                    Partial CE + Tags + Common size     & 0.7900     & 0.7214 \\
                    Partial CE + Tags + Individual size & \textbf{0.8604}     & \textbf{0.8298}  \\ % Presize_upper
                    Fully supervised                    & 0.8999     & 0.8911          \\
                    \hline
                \end{tabular}
                \label{table:dices_spine_prostate}
            \end{table}

            \rred{For the prostate dataset, one can observe that common bounds still improve the results obtained with tags ($+3\%$), but the difference is much smaller than the case of vertebral-body segmentation. Using individual bounds increases the DSC value by $10\%$, reaching $0.8298$, a behaviour similar to what we observed earlier for the other datasets. Nevertheless, in this case, the gap between full and weak supervision with individual bounds constraints is larger than what we obtained for the other datasets.}

            %\rred{ This can be partly explained by the fact that the prostate has much more variability in terms of size and shape (compared to the spine) ; the images are less structured.} \JD{For the individual bounds it shouldn't matter the variability on size..}

        \subsection{Qualitative results:}
        \label{sssec:qual}

\begin{figure}[h!]
                \centering
                \includegraphics[width=\textwidth]{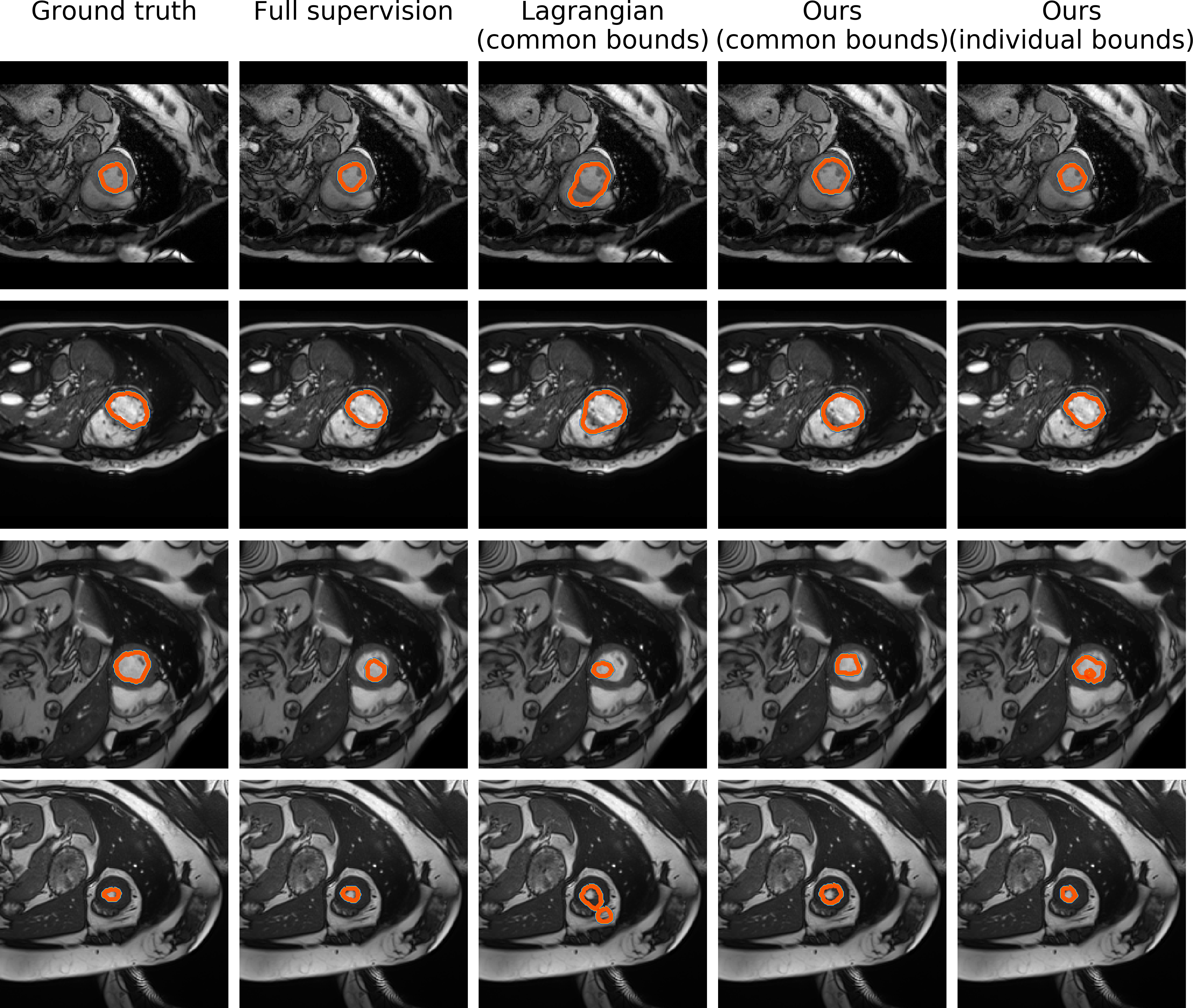}
                \caption{\rred{Qualitative comparison of the different methods using examples from the LV dataset. Each column depicts segmentations obtained by different methods, whereas each row represents a 2D slice from different scans (Best viewed in colors).}}
                \label{fig:cherrypick}
            \end{figure}

\begin{figure}[h!]
                \centering
                \includegraphics[width=1\textwidth]{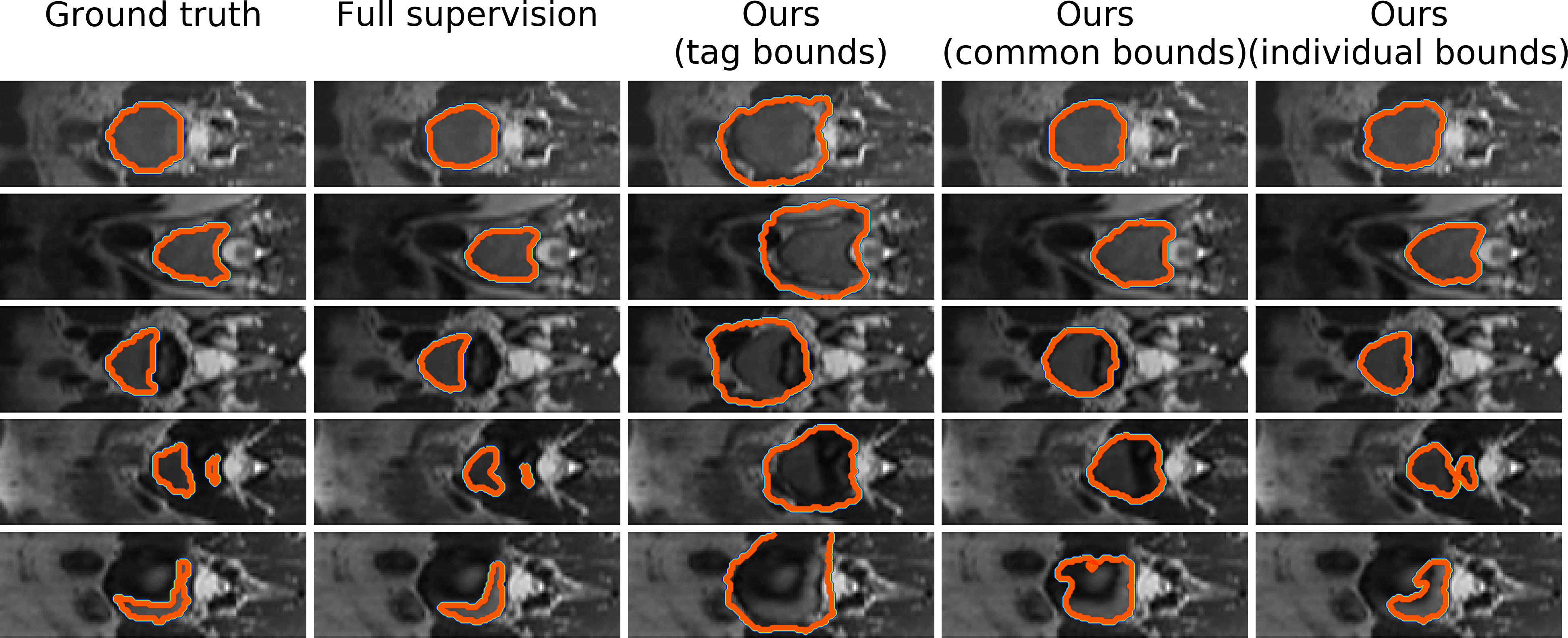}
                \caption{\rred{Qualitative comparison using examples from the VB dataset. Each column depicts segmentations obtained by different levels of supervision, whereas each row represents a 2D slice from different scans (Best viewed in colors).}}
                \label{fig:spine_results}
            \end{figure}

\begin{figure}[h!]
                \centering
                \includegraphics[width=1\textwidth]{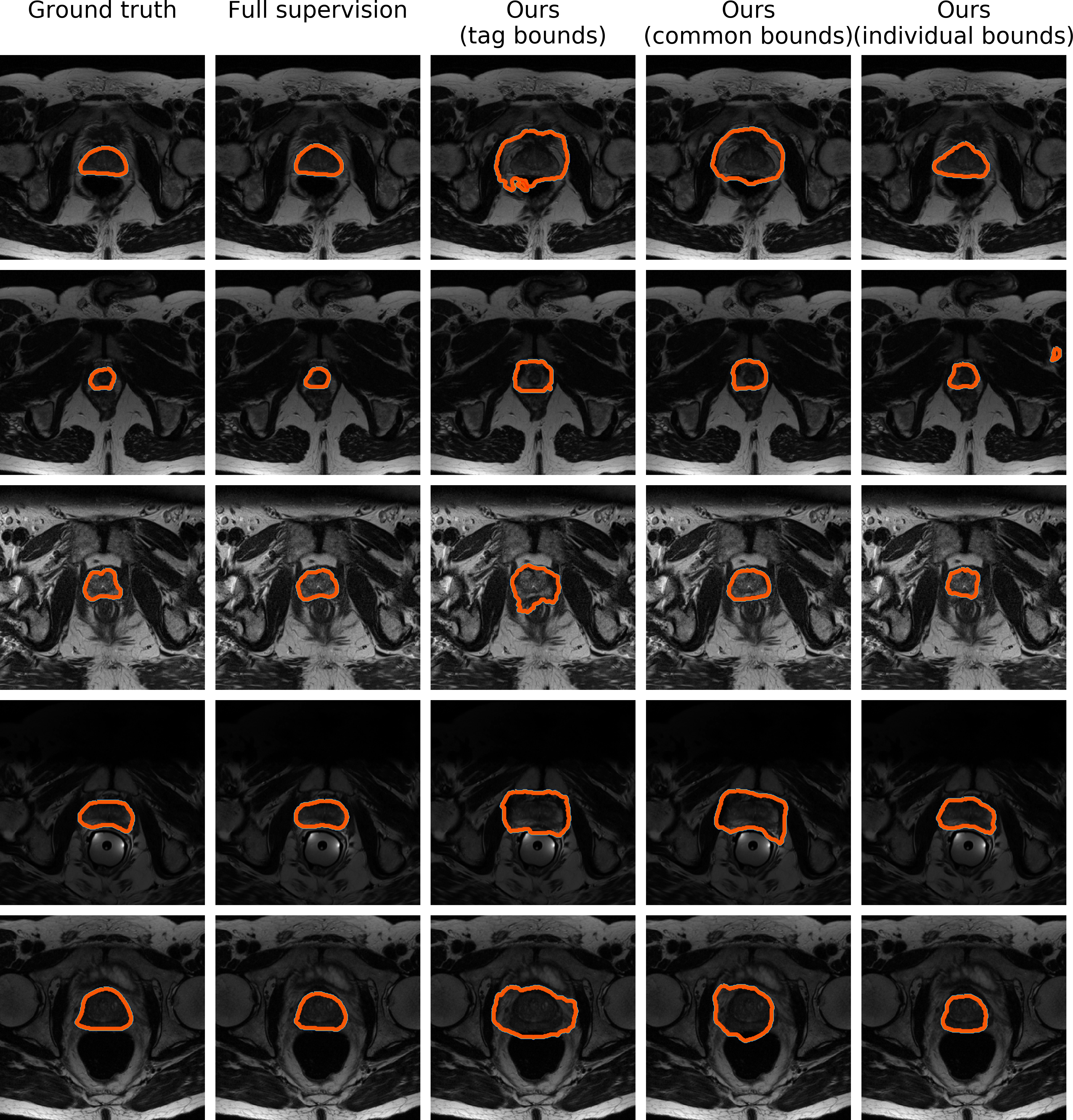}
                \caption{\bblue{Qualitative comparison of the different levels of supervision. Each row represents a 2D slice from different scans. (Best viewed in colors)}}
                \label{fig:prostate_results}
            \end{figure}

            To gain some intuition on different learning strategies and their impact on the segmentation, we visualize some results \rred{sampled from the validation sets in Fig. \ref{fig:cherrypick}, \ref{fig:spine_results} and \ref{fig:prostate_results} for LV, VB and prostate, respectively.}

            \rred{\paragraph{LV segmentation task} We compare 4 methods to the ground truth: full supervision, Lagrangian proposals \cite{pathak2015constrained} with common bounds, direct loss with common bounds and direct loss with individual bounds. We can see that, for the easy cases containing regular shapes and visible borders, all methods obtain similar results. However, the methods employing common bounds can easily over-segment the object, especially when their size is considerably smaller; see for example the last row in Figure \ref{fig:cherrypick}. Since individual bounds are specific to each image, a model trained with these bounds will not suffer in such cases, as shown in the figure.}

            \rred{\paragraph{Vertebral-body segmentation task} In this case, we visualize the results of full supervision, tag bounds, common bounds and individual bounds. In line with results reported in Table \ref{table:dices_spine_prostate}, we can visually observe the gap in performances between each setting, which clearly highlights the impact of the different values of the bounds during the optimization process. Using only tags, the network learn to roughly locate the object. When size bounds are included as common size information, the network is able to somehow learn the boundaries, but only for object  shapes that are within the standard variability of a typical vertebral body shape. As it can be observed, the model fails to segment the unusual shapes (last three rows in Figure \ref{fig:spine_results}). Lastly, a network trained with individual sizes is able to better handle those cases, while still being imprecise on some regions. }

            \rred{\paragraph{Prostate segmentation task} As in the previous case, we depict the results of full supervision, tag bounds, common bounds and individual bounds. Both the tags and common bounds locate the object in a similar fashion, but both have difficulties finding a precise contour, typically over-segmenting the target region. This is easily explained by the variability of the organ and the very low contrast on some images. As shown in the last column, using individual bounds greatly improves the results.}

        \subsection{Sensitivity to the constraint boundaries:}
        \label{ssec:ablation_bounds}

            In this section, an ablation study is performed on the lower and upper bounds \rred{when using common bounds}, and investigate their effect on the performance \rred{on the vertebral-body segmentation task}. Results for different bounds are reported in Table \ref{table:bounds}. \rred{It can be observed that progressively increasing the value of the upper bound decreases the performance. For example, the DSC drops by nearly 12\% and 16\% when the upper bound is increased by a factor of 5 and 10, respectively.
           % , indicating that the performance is more sensitive to bound values when they are `closer' to the real target size. On the other hand,
            Decreasing the lower bound from 80 to 0 has a much smaller impact than the upper bound, with a constant drop of less than 1\%. These findings are aligned with visual predictions illustrated in Figure \ref{fig:spine_results}. While a network trained only with tag bounds tends to over-segment, adding an upper bound easily fixes the over-segmentation, correcting most of the mistakes. Nevertheless, for the same reason, i.e., over-segmentation, very few slices benefit from a lower bound.}

            \begin{table}[h!]
                \footnotesize
                \centering
                \caption{\bblue{Ablation study on the lower and upper bounds of the size constraint} \rred{using the vertebral body dataset.}}
                \label{table:bounds}
                \begin{tabular}{lccc}
                           & \multicolumn{2}{c}{\textbf{Bounds}}   & \textbf{Mean DSC} \\ \midrule
                    \textbf{Model}          & \multicolumn{1}{l}{Lower (a)} & \multicolumn{1}{l}{Upper (b)} & \multicolumn{1}{l}{}  \\ \midrule
                    Weak Sup. w/ direct loss & $0.9\tau_Y$  & $1.1\tau_Y$   & 0.8604 \\
                    \midrule \midrule
                    Weak Sup. w/ direct loss & 80  & 1100   & 0.7900 \\
                    % Weak Sup. w/ direct loss & 80  & 2000  & 0.6260 \\
                    Weak Sup. w/ direct loss & 80  & 5000  & 0.6704 \\
                    Weak Sup. w/ direct loss & 80  & 10000 & 0.6349 \\
                    \midrule
                    Weak Sup. w/ direct loss & 0  & 1100  & 0.7820 \\
                    Weak Sup. w/ direct loss & 0  & 5000   & 0.6694 \\
                    Weak Sup. w/ direct loss & 0  & 10000  & 0.6255 \\
                    \midrule \midrule
                    Weak Sup. w/ direct loss & 0  & 65536  & 0.5597 \\
                    \bottomrule
                \end{tabular}
            \end{table}

        \subsection{Efficiency:}
        \label{sssec:times}

            In this section, we compare the several learning approaches in terms of efficiency (Table \ref{table:times}). Both the weakly supervised partial cross-entropy and the fully supervised model need to compute only one loss per pass. This is reflected in the lowest training times reported in the table. Including the size loss does not add to the computational time, as can be seen in these results. As expected, the iterative process introduced by \cite{pathak2015constrained} at each forward pass adds a \rred{significant} overhead during training. To generate their synthetic ground truth, they need to optimize the Lagrangian function with respect to its dual variables (Lagrange multipliers of the constraints), which requires alternating between training a CNN and Lagrangian-dual optimization. Even in the simplest optimization case (with only one constraint), where optimization over the dual variable converges rapidly, \rred{their method remains two times slower than ours. Without the early stopping criteria that we introduced, the overhead is much worse with a six-fold slowdown. In addition, their method also slows down when more constraints are added. This is particularly significant when there is many classes to constrain/supervise.}

            \rred{Generating the proposals at each iteration also makes it much more difficult to build an efficient implementation for larger batch sizes. One either needs to generate them one by one (so the overhead grows linearly with the batch size) or try to perform it in parallel. However, due to the nature of GPU design, the parallel Lagrangian optimizations will slow each other down, meaning that there may be limited improvements over a sequential generation. In some cases it may be faster to perform it on CPU (where the cores can truly perform independent tasks in parallel), at the cost of slow transfers between GPU and CPU. The optimal strategy would depend on the batch size and the host machine, especially its available GPU, number of CPU cores and bus frequency.}

            \begin{table}[h!]
                \centering
                \caption{Training times for the diverse supervised learning strategies \rred{with a batch size of 1}, using tags and size constraints.}
                \begin{tabular}{l|c}  \hline
                    Method                             & Training time (ms/batch) \\ \hline
                    Partial CE                         & 112              \\
                    \rred{Direct loss (1 bound)}         & 113              \\
                    \rred{Direct loss (2 bounds)}         & 113              \\
                    \rred{Lagrangian proposals (1 bound)} & 610              \\
                    \rred{Lagrangian proposals (2 bounds)} & 675              \\
                    \rred{Lagrangian proposals (1 bound), w/ early stop} & 221              \\
                    \rred{Lagrangian proposals (2 bounds), w/ early stop} & 220              \\
                    Fully supervised                   & 112              \\ \hline
                \end{tabular}

                \label{table:times}
            \end{table}

%------------------------------
    \section{Discussion}

        \bblue{We have presented a method to train deep CNNs with linear constraints in weakly supervised segmentation. To this end, we introduce a differentiable term, which enforces inequality constraints directly in the loss function, avoiding expensive Lagrangian dual iterates and proposal generation.}

        \bblue{Results have demonstrated that leveraging the power of weakly annotated data with the proposed direct size loss is highly beneficial, particularly when limited full annotated data is available. This could be explained by the fact that the network is already trained properly when a large fully annotated training set is available, which is in line with the values reported in Table \ref{table:hybrid}. Similar findings were reported in \cite{bai2017semi,zhou2018semi}, where authors exhibited an increased of performance when including non-annotated images in a semi-supervised setting. This suggests that including more unlabelled or weakly labelled data can potentially lead to significantly improvements in performance. }

        \rred{Findings from experiments across different segmentation tasks indicate that highly competitive performance can be obtained with a rough estimation of the target size. This is especially the case on well structured problems where the size and/or shape of the object remains consistent across subjects. If more precise size bounds are provided, the proposed approach is able to reach performances close to full supervision, even when the size and shape variability across subjects is large. For difficult tasks, where the gap between our approach and full supervision is larger, such as prostate segmentation, including an unsupervised regularization loss \cite{ncloss:cvpr18,tang2018regularized} to encourage pairwise consistencies between pixels may boost the performance of the proposed strategy. A noteworthy point is the robustness of our method to the weak-label generation. While the weak labels were generated from a ground-truth erosion for the first dataset, with seeds always in the center of the target region, they were randomly generated and placed  for the other two datasets. Thus, the results showed consistency in the behaviour of the different methods, regardless of the strategy used.}

        \rred{Even though the proposed method has been shown to provide good generalization capabilities across three different applications, the segmentation of images with severe abnormalities, whose sizes largely differ from those seen in the training set, has not been assessed. Nevertheless, the ablation study performed on the values of the size bounds, and the results obtained with common bound sizes suggest that the proposed approach may perform satisfactorily in the presence of these severe abnormalities, by simply increasing the upper bound value. In addition, if a greater `precise' estimation of the abnormality size is given, our proposed loss may improve segmentation performance, as demonstrated by the results achieved by the individual bounds strategy. It is important to note that, even in the case of full supervision, if a new testing image contains a severe abnormality much larger than the objects seen during the training phase, the network will likely to poorly segment the region of interest.}

        \bblue{Our framework can be easily extended to other non-linear (fractional) constraints, e.g., invariant shape moments \cite{Klodt2011} or other statistics such as the mean of intensities within the target regions \cite{Lim2014}. For instance, a normalized (scale invariant) shape moment of a target region can be directly expressed
        in term of network outputs using the following general fractional form:
        \begin{equation}
            \label{general-fractional-term}
            F_S = \frac{\sum_{p \in \Omega}f_p S_p}{\sum_{p \in \Omega}S_p}
        \end{equation}
        where $f_p$ is a unary potential expressed in term of exponents of pixel/voxel coordinates. For example, the coordinates of the center of mass of the target region are particular cases of \eqref{general-fractional-term} and correspond to first-order scale-invariant shape moments. In this case, potentials $f_p$ correspond to pixel coordinates. Now, assume a weak-supervision scenario in which we have a rough localization of the centroid of the target region. In this case, instead of a constraint on size representation $V_S$ as in Eq. \eqref{eq:lossForward1}, one can use a cue on centroid as follows: $a \leq F_S \leq b$. This can be embedded as a direct loss using differentiable penalty ${\cal C}(F_S)$. Of course, here, $F_S$ is a non-linear fractional term unlike region size. Therefore, in future work, it would be interesting to examine the behaviour of such fractional terms for constraining deep CNNs with a penalty approach.
        Finally, it is worth noting that the general form in Eq. \eqref{general-fractional-term} is not confined to shape moments. For instance, the image (intensity) statistics within the target region, such as the mean\footnote{Notice that the mean of intensity within the target region can be represented with network output using general form \eqref{general-fractional-term}, with $f_p$ corresponding to the intensity of pixel $p$}, follow the same general form in \eqref{general-fractional-term}. Therefore, a similar approach could be used in cases where we have prior knowledge on such image statistics.}

        \bblue{Our direct penalty-based approach for inequality constraints yields a considerable increase in performance with respect to to Lagrangian-dual optimization \cite{pathak2015constrained}, \rred{while being  faster and more stable}.
        We hypothesize that this is due, in part, to the
        {\em interplay} between stochastic optimization (e.g., stochastic gradient descent) for the primal and the iterates/projections for the Lagrangian dual\footnote{In fact, a similar hypothesis was made in \cite{Marquez-Neila2017} to explain the negative results of Lagrangian optimization in the case of equality constraints.}. Such dual iterates/projections are basic (non-stochastic) gradient methods for handling the constraints. Basic gradient methods have well-known issues with deep networks, e.g., they are sensitive to the learning rate and prone to weak local minima. Therefore, the dual part in Lagrangian optimization might obstruct the practical and theoretical benefits of stochastic optimization (e.g., speed and strong generalization performance), which are widely established for unconstrained deep network losses \cite{Hardt2016}. Our penalty-based approach transforms a constrained problem into an unconstrained loss, thereby handling the constraints fully within stochastic optimization and avoiding completely the dual steps. While penalty-based approaches do not guarantee constraint satisfaction, our work showed that they can be extremely useful in the context of constrained CNN segmentation.}

    \section{Conclusion}

        \rred{In this paper, a novel loss function is present for weakly supervised image segmentation, which, despite its simplicity, performs significantly better than Lagrangian optimization for this task. We achieve results close to full supervision by annotating only a small fraction of the pixels, across three different tasks, and with negligible computation overhead}. While our experiments focused on basic linear constraints such as the target-region size and image tags, our direct constrained-CNN loss can be easily extended to other non-linear constraints, e.g., invariant shape moments \cite{Klodt2011} or other region statistics \cite{Lim2014}. Therefore, it has the potential to close the gap between weakly and fully supervised learning in semantic medical image segmentation.

    \subsubsection*{Acknowledgments}

        This work is supported by the National Science and Engineering Research Council of Canada (NSERC), discovery grant program, and by the ETS Research Chair on Artificial Intelligence in Medical Imaging.

    \section*{References}
    \small

    \bibliographystyle{elsarticle-num}
    \bibliography{refs}
\end{document}